\documentclass[letterpaper, 10pt, conference]{ieeeconf}

\IEEEoverridecommandlockouts                              
\usepackage{graphicx}
\usepackage{amsmath,amssymb,amsfonts,bm}
\usepackage{booktabs}
\usepackage{multirow}
\usepackage{makecell}
\usepackage{array}
\usepackage{url}
\usepackage{cite}
\usepackage[dvipsnames]{xcolor}
\usepackage{capt-of}
\usepackage{needspace}

\graphicspath{{fig/}}

\newcommand{\method}{LightVLN}

\title{\LARGE \bf
\method: Efficient Aerial Vision-and-Language Navigation\\
with Compact Memory and History-Guided Local Aggregation
}

\author{\makebox[\textwidth][c]{Yiming Zhao, Tianshun Li, Jingle He, Ruonan Chai, and Xinhu Zheng$^{*}$}%
\makebox[0pt][r]{\makebox[\textwidth][c]{\raisebox{-13pt}[0pt][0pt]{\small The Hong Kong University of Science and Technology (Guangzhou), Guangzhou, China}}}%
\makebox[0pt][r]{\makebox[\textwidth][c]{\raisebox{-24pt}[0pt][0pt]{\footnotesize $^{*}$Corresponding author.}}}}

\begin{document}
\IEEEaftertitletext{%
  \begin{center}
    \includegraphics[width=0.9\textwidth]{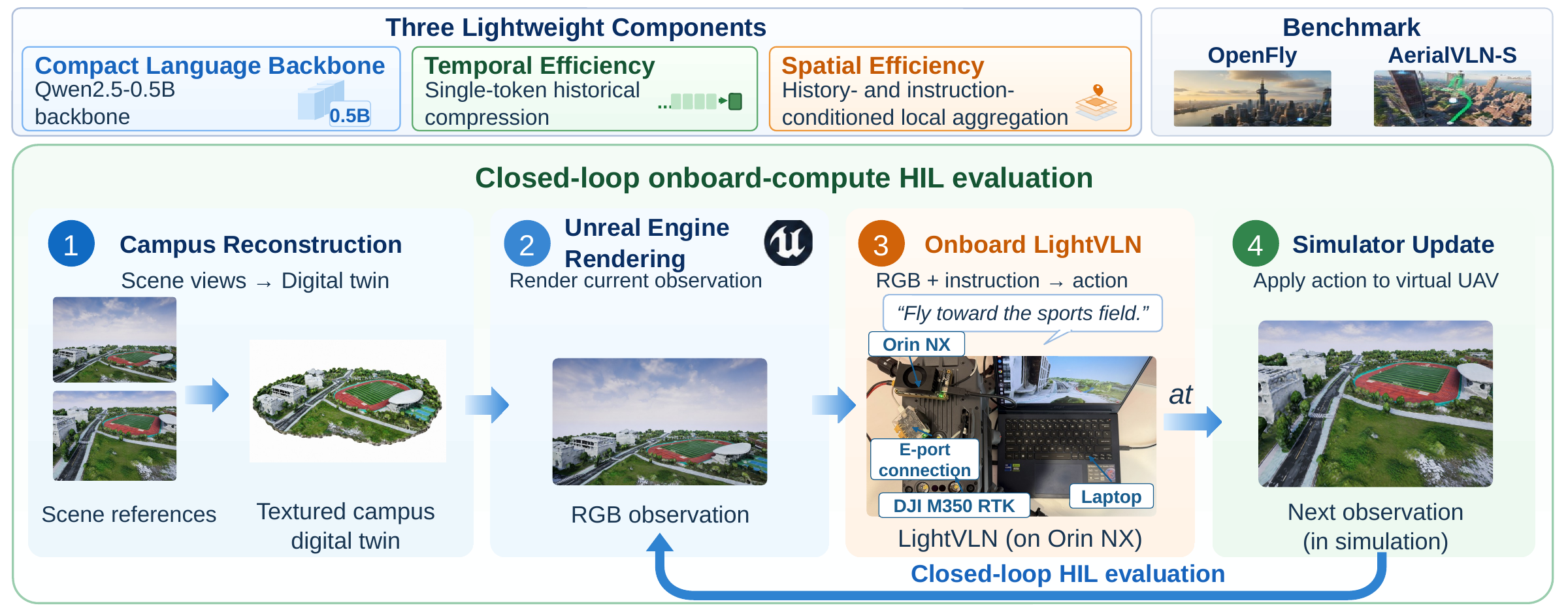}
    \captionof{figure}{\method{}: lightweight components, benchmark evaluation, and closed-loop onboard-compute HIL.
    In the Unreal Engine campus environment, AirSim acquires images,
    \method{} predicts actions on Orin NX, and AirSim applies the actions
    to the virtual UAV to obtain the next observation.}
    \label{fig:overview}
  \end{center}}
\maketitle
\thispagestyle{empty}
\pagestyle{empty}

\begin{abstract}
Aerial vision-and-language navigation (VLN) enables unmanned aerial
vehicles to execute long-horizon natural-language instructions from
visual observations in complex three-dimensional environments.
However, recent aerial VLN models often rely on large-scale
vision-language backbones and dense visual histories, imposing
substantial computation and memory costs that hinder onboard deployment.
We propose \method{}, a lightweight history-aware aerial VLN framework
that combines a compact 0.5B language backbone with compact representations
of both historical and current observations.
\method{} compresses each
historical frame into a single token using visual features already computed by the policy. 
It further introduces history-
and instruction-conditioned local aggregation to reduce the current
observation from 256 to 32 visual tokens while preserving
navigation-relevant spatial information. With up to 16 historical frames,
the policy uses at most 48 observation-derived tokens. 
On the public OpenFly dataset, \method{} achieves 50.93\% Test-Seen
and 36.14\% Test-Unseen success rates (SR), outperforming the evaluated
7B language-backbone baselines on most reported metrics. 
It also achieves 25.83\% SR on AerialVLN-S Val-Seen.
In a reconstructed unseen campus, we deploy \method{} on a DJI M350 RTK
with an external Jetson Orin NX 16\,GB for closed-loop onboard-compute
real-to-sim hardware-in-the-loop (HIL) evaluation, achieving 14.61\,Hz
model inference and 11.13\,Hz end-to-end decision updates.
These results demonstrate the effectiveness and efficiency of LightVLN for aerial navigation.
\end{abstract}

\section{Introduction}
Aerial vision-and-language navigation (VLN) enables unmanned aerial vehicles (UAVs) to follow long-horizon natural-language instructions from visual observations in complex three-dimensional environments~\cite{liu2023aerialvln}. It requires the UAV to connect instructions with visual observations and predict a sequence of actions toward the navigation goal. The AerialVLN benchmark established a foundation for this task through simulated aerial environments, annotated navigation trajectories, and a cross-modal alignment baseline~\cite{liu2023aerialvln}. Subsequent platforms expanded the realism and scale of aerial navigation data, supporting the development of more advanced navigation models~\cite{wang2025openuav,gao2025openfly}. Recent methods further improve spatial reasoning and long-horizon decision making through explicit spatial representations, semantic subgoals, grounded targets, and learned multimodal policies~\cite{gao2024stmr,citynavagent2025,lookasidevln2026,xu2026fly0,aeroact2025}.

Aerial navigation models have increasingly adopted larger language backbones and richer visual histories. Video-based navigation models~\cite{zhang2024navid,navila2025} and aerial foundation-model policies~\cite{gao2025openfly,aeroact2025,deng2026dreamfly} use large pretrained multimodal backbones for visual grounding and action prediction. Meanwhile, long-horizon aerial navigation has motivated retaining information across observation sequences through selected-frame histories~\cite{gao2025openfly,aeroact2025}, structured spatiotemporal memory~\cite{ding2026hett,jiang2025longfly}, and relevance-aware or causal history integration~\cite{ma2026s2d,deng2026dreamfly}. These mechanisms provide access to historical visual information for subsequent navigation decisions. However, the use of larger language backbones and richer visual histories also brings substantial computation and memory demands at inference~\cite{zhang2024navid,wu2025vlaan}.

Deploying aerial navigation models on resource-constrained UAV platforms remains challenging~\cite{wu2025vlaan}. In particular, large backbones and extensive visual context increase computation and memory demands at inference. Recent deployment-oriented systems improve efficiency through optimized inference pipelines, lightweight action generation, shared perception, and fast--slow or asynchronous execution~\cite{wu2025vlaan,zhu2026fsdvln,zheng2026onfly}. Complementary to these deployment-oriented techniques, we target a compact language backbone and compact representations of both historical and current observations. Lightweight aerial VLN therefore also requires reducing the intrinsic resource demands of the navigation policy while preserving its navigation capability.

Designing such a policy poses two challenges: constructing low-cost historical memory that remains useful for action prediction, and compressing the current observation while preserving navigation-relevant spatial information.
First, for historical memory, existing aerial VLN methods typically control temporal redundancy and computation through keyframe selection or compact history representations~\cite{gao2025openfly,aeroact2025,jiang2025longfly}. These strategies reduce memory cost by limiting the number of retained observations or compressing their representations.
NativeMEM~\cite{wang2026nativemem} addresses this trade-off by compressing each historical frame into a single action-supervised token, enabling substantially longer and denser visual histories. It constructs memory through a separately optimized visual-encoding path initialized from the policy's vision encoder. 
Second, for current-view compression, existing methods reduce visual redundancy through learned-query bottlenecks, regional aggregation, or language-conditioned token selection~\cite{li2023blip2,li2024tokenpacker,zhang2024navid,song2026evocomp}. These methods effectively compress the current observation but do not exploit navigation history. DreamFly~\cite{deng2026dreamfly} incorporates historical context to enhance current-view representations through causal memory, but does not target lightweight visual token compression or low-cost memory construction.
These observations motivate an integrated lightweight design that combines low-cost historical memory with compact current-view representations.

To address these challenges, we propose \method{}, a lightweight history-aware aerial VLN framework that combines a compact 0.5B language backbone with compact representations of both historical and current observations (Fig.~\ref{fig:overview}).
Each historical frame is compressed into a single token using visual features already extracted by the policy. 
For the current observation, history- and instruction-conditioned local aggregation reduces 256 visual tokens to 32 while preserving navigation-relevant spatial information. 
With up to 16 historical frames, the resulting visual context contains at most 48 observation-derived tokens. 
We evaluate \method{} on the public OpenFly splits~\cite{gao2025openfly} and AerialVLN-S~\cite{liu2023aerialvln} and conduct extensive same-backbone ablations on OpenFly. We further reconstruct a campus environment and construct a 500-episode unseen evaluation set, where \method{} is deployed on a DJI M350 RTK with an external Jetson Orin NX 16\,GB for closed-loop onboard-compute real-to-sim HIL evaluation.

The main contributions are:
\begin{enumerate}
  \item We propose \method{}, a lightweight history-aware aerial VLN
        framework that combines a compact 0.5B language backbone with
        low-cost historical memory and compact current-view representations.
        With at most 48 observation-derived tokens, \method{} provides a compact visual interface for aerial VLN with a substantially smaller language backbone.

  \item We develop two complementary lightweight designs:
        single-token historical compression that reuses visual features
        already extracted by the policy, and history- and
        instruction-conditioned local aggregation that reduces the current
        observation from 256 to 32 visual tokens while preserving spatial
        information relevant to navigation.

  \item We evaluate \method{} on OpenFly and AerialVLN-S and conduct
        extensive same-backbone ablations on OpenFly. \method{} achieves 50.93\% Test-Seen and
        36.14\% Test-Unseen SR on public OpenFly, outperforming the
        evaluated 7B baselines on most reported OpenFly metrics, and achieves 25.83\% SR on
        AerialVLN-S Val-Seen. We further conduct closed-loop onboard-compute
        real-to-sim HIL evaluation in a reconstructed unseen campus, where \method{}
        reaches 14.61\,Hz model inference and 11.13\,Hz end-to-end decision
        updates on Jetson Orin NX 16\,GB.
\end{enumerate}

\begin{figure*}[!t]
  \centering
  \includegraphics[width=0.9\textwidth]{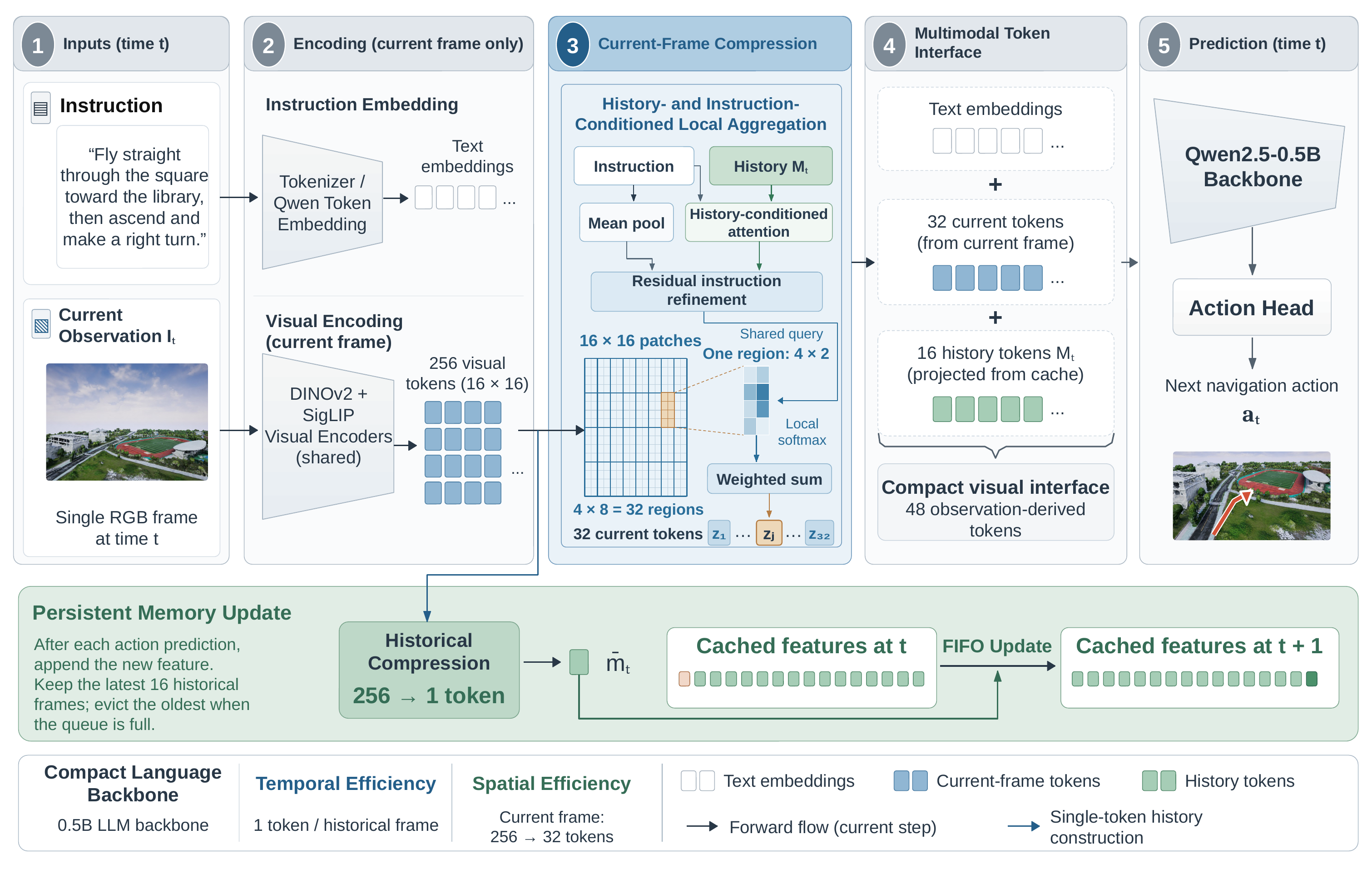}
  \caption{\method{} architecture. The main inference pipeline consists
  of inputs, encoding, current-observation compression, multimodal token
  interface, and action prediction. A persistent memory branch constructs
  one compact representation for each observed frame and updates the
  historical memory after action prediction. At $H=16$, the policy
  receives at most 48 observation-derived tokens together with the
  instruction.}
  \label{fig:architecture}
\end{figure*}

\section{Related Work}
\subsection{Aerial Vision-and-Language Navigation}
Existing aerial VLN methods broadly include structured reasoning and
learned multimodal navigation approaches.
Structured methods introduce explicit spatial or semantic representations
to support navigation reasoning. STMR~\cite{gao2024stmr} constructs
spatial representations for language-guided reasoning, while
CityNavAgent~\cite{citynavagent2025} performs hierarchical semantic
planning with global memory. LookasideVLN~\cite{lookasidevln2026}
incorporates direction-aware reasoning, and Fly0~\cite{xu2026fly0}
uses persistent metric anchors for zero-shot navigation.
Learned multimodal approaches directly predict navigation decisions from
visual observations and language instructions.
AerialVLN~\cite{liu2023aerialvln} establishes an early cross-modal
navigation baseline, while OpenFly-Agent~\cite{gao2025openfly} scales
aerial VLN with pretrained multimodal models and historical observations.
AeroAct~\cite{aeroact2025} further integrates spatial, temporal, and
embodied reasoning within a unified navigation model.

\subsection{Efficient Aerial VLN and Onboard Deployment}
Existing work improves the efficiency of aerial VLN through lightweight
inference, decoupled decision making, and computation reuse.
VLA-AN~\cite{wu2025vlaan} combines lightweight action generation with
an optimized inference pipeline for onboard deployment.
FSD-VLN~\cite{zhu2026fsdvln} adopts a fast--slow dual-system architecture,
separating high-level semantic reasoning from frequent navigation updates.
OnFly~\cite{zheng2026onfly} shares perception features and reuses memory
across decision steps to reduce redundant computation and support
real-time onboard execution.

\subsection{Compact Historical and Current Visual Representations}
Existing work reduces visual redundancy from two complementary aspects:
historical representation and current-view compression.
For historical representation, OpenFly-Agent~\cite{gao2025openfly}
and AeroAct~\cite{aeroact2025} retain selected historical observations
to provide temporal context while controlling the number of stored views.
HETT~\cite{ding2026hett} organizes historical information through
history-enhanced spatial representations, while LongFly~\cite{jiang2025longfly}
integrates compact spatiotemporal context for long-horizon navigation.
S2D~\cite{ma2026s2d} further emphasizes relevance-aware historical
information for decision making.
Beyond aerial VLN, NativeMEM~\cite{wang2026nativemem} compresses each
historical frame into a single action-supervised token, enabling much
denser temporal memory. It constructs memory through a separately
optimized visual-encoding path initialized from the policy's vision
encoder. Our memory directly reuses the policy's existing patch features.

For current-view compression, BLIP-2~\cite{li2023blip2} uses a learned
query bottleneck to extract a compact set of visual features, while
TokenPacker~\cite{li2024tokenpacker} aggregates local visual tokens to
preserve spatial structure with fewer tokens.
NaVid~\cite{zhang2024navid} introduces language-conditioned visual
compression for navigation, and EvoComp~\cite{song2026evocomp} selects
task-relevant visual tokens through semantic guidance.
AeroAct~\cite{aeroact2025} also employs regional visual aggregation in
aerial navigation.
These methods mainly focus on compressing the current observation itself,
whereas navigation history is usually handled as a separate source of
context.
DreamFly~\cite{deng2026dreamfly} instead uses causal memory to enhance
current-view representations with historical information.
We use compact historical memory to condition fixed-region aggregation
of the current view, while constructing that memory from visual
features already computed by the policy.

\section{\method{}}
\label{sec:lightvln}

\method{} combines a compact 0.5B language backbone with compact
historical and current visual representations. At decision step $t$,
it predicts $a_t=\pi_\theta(I_t,\mathcal L,M_t)$ from the current RGB
observation $I_t$, instruction $\mathcal L$, and historical tokens
$M_t=[m_{t-H_t},\ldots,m_{t-1}]$, where $0\le H_t\le H$ and $H=16$
is the default history capacity.

Figure~\ref{fig:architecture} illustrates the five-stage inference
pipeline: inputs, encoding, current-observation compression, multimodal
token interface, and action prediction. In parallel, a persistent memory
branch constructs a compact representation of each observed frame and
updates the historical memory across decision steps.

\subsection{Inputs and Encoding}
\label{sec:encoding}

At each decision step, the model receives the current RGB observation
$I_t$, the instruction $\mathcal L$, and the historical memory $M_t$
constructed from previous observations.

For visual encoding, frozen DINOv2~\cite{oquab2024dinov2} and
SigLIP~\cite{zhai2023siglip} encoders produce aligned $16\times16$
patch grids from $I_t$. Features at corresponding spatial locations are
concatenated into $X_t\in\mathbb R^{256\times d_r}$, where $d_r=2176$.
An MLP projector maps them to
$F_t=P_V(X_t)\in\mathbb R^{256\times d}$,
where $d=896$ is the language-model width.

For instruction encoding, Qwen's tokenizer and token-embedding layer
~\cite{qwen2025qwen25} produce
$E_{\mathcal L}=[e_1,\ldots,e_L]\in\mathbb R^{L\times d}$.
These encoded representations are subsequently used by the
single-token historical compression, history- and instruction-conditioned
current-observation aggregation, multimodal token interface, and action
prediction modules.

\subsection{Single-Token Historical Compression}
\label{sec:single-token-history}

Following the single-token memory design of
NativeMEM~\cite{wang2026nativemem}, we construct one token per
historical frame using visual features already extracted by the policy.
A learned input-independent query $q_0^N\in\mathbb R^{d_r}$ pools the
current-frame features $X_t$ using multi-head attention (MHA):
\begin{equation}
\bar m_t=\operatorname{MHA}(q_0^N,X_t,X_t),\qquad
m_t=\operatorname{LN}(P_M(\bar m_t))\in\mathbb R^d.
\label{eq:memory}
\end{equation}
The tokenizer uses eight attention heads with internal width 2176, reducing the $256\times2176$ visual features
to a single $2176$-dimensional representation $\bar m_t$.
The memory projector $P_M$ maps the cached representations to the
language-model width $d=896$. At decision step $t$, the cached features
from previous frames are projected and normalized to form
$M_t=[m_{t-H_t},\ldots,m_{t-1}]$.

\textbf{Persistent memory update.}
After action prediction, the current representation $\bar m_t$ is
appended to a first-in, first-out (FIFO) cache. The cache retains the
most recent $H$ frame representations and discards the oldest one when
its capacity is exceeded. The updated cache is used to construct
$M_{t+1}$ for the next decision step. This design requires no additional memory encoder or online re-encoding of historical images.

\subsection{History- and Instruction-Conditioned Local Aggregation}
\label{sec:spatial}

The current observation contains 256 visual tokens in $F_t$. To obtain
a compact representation while retaining spatial coverage, we use the
historical memory and instruction to guide local aggregation.

\textbf{History-conditioned instruction representation.}
We first mean-pool the valid instruction embeddings to obtain a global
instruction representation $e_G$. The historical tokens are zero-left-padded to length $H$ and summarized
along the temporal dimension by a shared $H\!\to\!32\!\to\!1$ MLP
with GELU activation:
\begin{equation}
h_t=\mathcal T_\phi\bigl(\operatorname{PadLeft}(M_t)\bigr),
\label{eq:history-summary}
\end{equation}
where $\mathcal T_\phi$ is applied independently to each feature channel,
yielding $h_t\in\mathbb R^d$.

We then use the history representation to select instruction content
relevant to the current navigation state:
\begin{equation}
e_{\mathrm{sel}}=
\operatorname{Attn}\left(W_q h_t,\,E_{\mathcal L}W_k^\top,\,E_{\mathcal L}\right),
\label{eq:history-instruction}
\end{equation}
where $\operatorname{Attn}$ denotes single-head scaled dot-product attention
and $e_{\mathrm{sel}}\in\mathbb R^d$ is the history-conditioned instruction
representation. When no historical observation is available,
we set $e_{\mathrm{sel}}=e_G$.

To retain the global instruction semantics while incorporating
history-conditioned information, we combine the two representations as
\begin{equation}
e_{\mathrm{ref}}=e_G+\gamma(e_{\mathrm{sel}}-e_G),
\label{eq:refinement}
\end{equation}
where $\gamma$ is a learnable scalar initialized to zero. The resulting
$e_{\mathrm{ref}}$ is used to guide current-view aggregation.

\textbf{Local aggregation.}
To preserve spatial coverage under a compact token budget, we partition
the $16\times16$ current-view feature grid into 32 non-overlapping
$4\times2$ regions. The refined instruction representation is projected
into a shared visual query $q_V=W_Ve_{\mathrm{ref}}$. For each region,
we locally aggregate its eight visual tokens as
\begin{equation}
\begin{aligned}
\alpha_{t,r,j}&=
\operatorname{softmax}_{j\in\mathcal R_r}
\left(
\frac{f_{t,j}^{\top}q_V}{\sqrt d}
\right),\\
z_{t,r}&=
\sum_{j\in\mathcal R_r}\alpha_{t,r,j}f_{t,j},
\end{aligned}
\label{eq:local-aggregation}
\end{equation}
where $\mathcal R_r$ denotes the $r$-th local region. This produces one
token per region, reducing the current observation from 256 to 32 visual
tokens while maintaining fixed spatial coverage. The resulting tokens
form $Z_t\in\mathbb R^{32\times d}$.

\subsection{Multimodal Token Interface and Action Prediction}
\label{sec:prediction}

The current-view tokens $Z_t$, instruction embeddings
$E_{\mathcal L}$, and valid historical tokens $M_t$ are assembled into
the input sequence of Qwen2.5-0.5B. A learned MEM-BOS embedding marks
the beginning of the historical block, and a learnable 896-dimensional
ACT query is appended at the end:
\begin{equation}
\mathcal S_t=
[\mathrm{BOS},Z_t,E_{\mathcal L},
\text{MEM-BOS},M_t,\mathrm{ACT}].
\label{eq:sequence}
\end{equation}
Only valid historical tokens are included and are ordered from oldest
to newest. Under causal attention, the final ACT query attends to all
preceding valid tokens. At $H=16$, the current and historical visual
blocks contain at most $32+H_t\le48$ observation-derived tokens,
excluding the instruction and special tokens.

Let $h_t^{\mathrm{ACT}}\in\mathbb R^{896}$ denote the final hidden state at the ACT position. A lightweight MLP action head with one hidden layer and LayerNorm maps this representation to eight action logits, followed by softmax for the action distribution. 

\subsection{Two-Stage Training}
\label{sec:two-stage-training}

We first obtain a well-trained single-frame Base policy with the
32-token current-view representation but without historical memory.
Starting from this Base policy, we adopt a two-stage training strategy
to learn stable single-token historical representations. All stages use
cross-entropy supervision over the demonstrated actions.

\textbf{Stage 1: memory representation learning.}
We freeze the well-trained Base policy and optimize only the historical
tokenizer, memory projector, and MEM-BOS embedding. The 32-token
current-frame pathway is retained, while the compressed historical tokens
are introduced through the multimodal token interface in
Eq.~\eqref{eq:sequence}. After Stage~1, the historical tokenizer is frozen
and used to generate the historical features cached for Stage~2.

\textbf{Stage 2: memory-augmented policy training.}
We train the final policy using the cached historical features. The
historical tokenizer and visual encoders remain frozen, while all
remaining trainable modules are jointly optimized.

\section{Experiments}
\subsection{Experimental Setup}
\label{sec:setup}
\textbf{Benchmarks and training inputs.}
Separate policies are trained on the released OpenFly and AerialVLN-S
training splits, using RGB images and instructions as inputs and
demonstrated actions as supervision. OpenFly training uses 100,221
episodes with 1,608,400 supervised step samples; AerialVLN-S training
uses 1,157,437 supervised step samples. OpenFly~\cite{gao2025openfly}
Test-Seen contains 1,392 public episodes from AirSim scenes 16, 18,
26, gz, sh, and 23, plus UE BigCity; Test-Unseen contains 404 episodes
from UE SmallCity. 
AerialVLN-S uses the original release, simulator, annotations, and
333-instruction Val-Seen split~\cite{liu2023aerialvln}.

\textbf{Campus unseen evaluation set.}
We reconstruct an anonymized campus environment from DJI Mavic 3E observations.
DJI Terra performs pose estimation, dense reconstruction, meshing,
and texture mapping; the metric textured model is imported into
UE/AirSim with OpenFly's categorical action interface and motion
settings~\cite{gao2025openfly}. The 500-episode campus unseen evaluation set
allocates balanced quotas across five sub-regions and 53 verified
goal landmarks. Collision-checked 3D routes have controlled
intra-/cross-region composition, length, and maneuver complexity.
Following the data-generation paradigm of OpenFly~\cite{gao2025openfly},
routes and navigation instructions are constructed from discrete actions
and key-view observations. The campus and its observations are reserved for
unseen evaluation of an OpenFly-trained checkpoint.

\textbf{Metrics and comparisons.}
We follow the official OpenFly and AerialVLN-S evaluation implementations.
We report navigation error (NE), success rate (SR), oracle success rate
(OSR), and success weighted by path length (SPL) on OpenFly. On
AerialVLN-S, we report NE, SR, OSR, and Success weighted by normalized
Dynamic Time Warping (SDTW).
Both benchmarks define successful navigation as stopping within 20\,m of
the goal, with collisions counted as failures~\cite{gao2025openfly,liu2023aerialvln}.
OSR measures whether the agent enters the 20\,m goal region at any point
during navigation. Campus evaluation follows the OpenFly protocol.
Table~\ref{tab:openfly} separates OpenFly-Agent's released scores from
our same-pipeline re-evaluation and groups six-scene AirSim-only
results separately from the public setting used by DreamFly
~\cite{deng2026dreamfly} and \method{}.
Table~\ref{tab:main} groups LAG, STMR, CityNavAgent, and LookasideVLN,
which use additional depth, panoramic observations, or odometry,
separately from monocular RGB methods
~\cite{liu2023aerialvln,gao2024stmr,citynavagent2025,lookasidevln2026}.
FLOPs and latency are profiled on an RTX 4090. Latency is averaged over 500 synchronized inferences after 20 warm-up iterations.

\textbf{Implementation.}
We initialize from MiniVLA's \texttt{prism-qwen25-0.5b-hf}
checkpoint~\cite{belkhale2024minivla}, using
Qwen2.5-0.5B-Instruct~\cite{qwen2025qwen25}, DINOv2 ViT-L/14 with
four register tokens, and SigLIP ViT-SO400M/14-224 at $224\times224$
resolution. 
The complete policy has approximately 1.276B parameters; 0.5B denotes the language backbone.
Offline training and inference use BF16 for most parameters and FP32
for numerically sensitive parameters. Both datasets use separately
trained eight-class action heads. For OpenFly, we exclude the two released
action categories that are unused in the released trajectories, resulting
in an eight-action policy space.

All stages use an effective batch size of 256. Base, Stage~1, and
Stage~2 are trained for 3, 1, and 4 epochs, respectively. The Stage~1
checkpoint is used to build the 1T cache, and the reported
model uses the Stage~2 epoch-4 checkpoint. Training uses two H100 80\,GB
GPUs and LoRA~\cite{hu2022lora} with rank $r=64$, $\alpha=16$, and
dropout 0.05. Learning rates are $10^{-4}$ for the historical tokenizer
and current-view visual projector, $5\times10^{-5}$ for the memory
projector and MEM-BOS, and $2\times10^{-4}$ for LoRA, current-view
aggregation, history-conditioning modules, and action modules, with
3\% linear warm-up and cosine decay. 

\textbf{Onboard setup and profiling.}
The closed-loop onboard-compute HIL platform couples a stationary, disarmed DJI M350 RTK carrying
an E-Port-connected Jetson Orin NX 16\,GB with a UE/AirSim host over
USB Ethernet. 
AirSim supplies rendered observations
and applies the Jetson's predicted actions through SetPose, confirming
the virtual-pose update before the next observation.
FLOPs and local latency cover complete policy inference.
All latency measurements use batch size 1. 
Onboard inference uses TensorRT 10.3.0
and TensorRT-LLM 0.12.0 with module-specific mixed precision and no
model offloading. 
The exported policy agrees with the offline reference
on all 500 action-check steps.
 Dense current and
\method{} use $H=16$, with the Jetson Orin NX configured in MAXN
power mode. Latency is averaged over 500 measured frames after
10 warm-up steps.
 End-to-end latency includes
image acquisition, communication, model inference, and virtual-pose
update/confirmation.

\subsection{Main Benchmark Results}
Among the compared methods in the public OpenFly setting, \method{}
achieves the best NE, SR, OSR, and SPL on Test-Seen and the best NE,
SR, and SPL on Test-Unseen (Table~\ref{tab:openfly}). It also leads all
four reported metrics on AerialVLN-S Val-Seen (Table~\ref{tab:main}).

\begin{table*}[t]
  \caption{OpenFly results. Scale: language-backbone size.}
  \label{tab:openfly}
  \centering
  \footnotesize
  \setlength{\tabcolsep}{2pt}
  \begin{tabular*}{\textwidth}{@{\extracolsep{\fill}}llcccccccc@{}}
    \toprule
    \multirow{2}{*}{Method} & \multirow{2}{*}{Scale} &
    \multicolumn{4}{c}{Test-Seen} &
    \multicolumn{4}{c}{Test-Unseen} \\
    \cmidrule(lr){3-6} \cmidrule(lr){7-10}
    & & NE$\downarrow$(m) & SR$\uparrow$(\%) & OSR$\uparrow$(\%) & SPL$\uparrow$(\%)
        & NE$\downarrow$(m) & SR$\uparrow$(\%) & OSR$\uparrow$(\%) & SPL$\uparrow$(\%) \\
    \midrule
    \multicolumn{10}{l}{\scriptsize\textit{AirSim-only setting: 6 Seen scenes, 1,210 episodes}} \\
    AeroAct$^\dagger$~\cite{aeroact2025} & 8B
      & 38.00 & 54.50 & 72.10 & 49.90 & -- & -- & -- & -- \\
    \midrule
    \multicolumn{10}{l}{\scriptsize\textit{Public dataset setting: 7 Seen scenes / 1,392 episodes, 1 Unseen scene / 404 episodes}} \\
    CMA$^\ddagger$~\cite{liu2023aerialvln} & --
      & 313.03 & 7.97 & 69.32 & 6.26
      & 230.05 & 5.69 & \textbf{73.02} & 3.92 \\
    AerialVLN$^\ddagger$~\cite{liu2023aerialvln} & --
      & 176.29 & 16.52 & 65.66 & 14.63
      & 161.19 & 9.65 & 68.07 & 7.93 \\
    OpenFly-Agent (reported)~\cite{gao2025openfly} & 7B
      & 118.72 & 28.76 & 63.10 & 18.71
      & 164.40 & 26.20 & 68.80 & 18.00 \\
    OpenFly-Agent (our re-eval.)$^\S$~\cite{gao2025openfly} & 7B
      & 107.39 & 25.07 & 56.68 & 22.54
      & 140.66 & 19.06 & 68.81 & 16.19 \\
    DreamFly~\cite{deng2026dreamfly} & 7B
      & 44.87 & 32.04 & 46.77 & 28.22
      & 45.29 & 29.46 & 46.78 & 23.54 \\
    \textbf{\method{} (Ours)} & \textbf{0.5B}
      & \textbf{40.25} & \textbf{50.93} & \textbf{71.41} & \textbf{44.68}
      & \textbf{42.22} & \textbf{36.14} & 62.13 & \textbf{30.45} \\
    \bottomrule
  \end{tabular*}
  \par\smallskip
  \begin{minipage}{\textwidth}
  \scriptsize
  $\dagger$ AeroAct trains only on AirSim and evaluates on six AirSim scenes.
  Public evaluation additionally includes one Seen and one Unseen scene.
  \par
  $\ddagger$ Results reported by DreamFly~\cite{deng2026dreamfly}.
  \quad
  $\S$ Released checkpoint re-evaluated on the same episodes and pipeline
  as \method{}.
  \end{minipage}
\end{table*}

\begin{table}[t]
  \caption{Performance on the AerialVLN-S Val-Seen.}
  \label{tab:main}
  \centering
  \footnotesize
  \setlength{\tabcolsep}{1.5pt}
  \begin{tabular*}{\columnwidth}{@{\extracolsep{\fill}}lcccc@{}}
    \toprule
    \multirow{2}{*}{Method} &
      \multicolumn{4}{c}{Val-Seen} \\
    \cmidrule(lr){2-5}
      & \makecell{NE$\downarrow$\\(m)} & \makecell{SR$\uparrow$\\(\%)}
        & \makecell{OSR$\uparrow$\\(\%)} & \makecell{SDTW$\uparrow$\\(\%)} \\
    \midrule
    LAG$^\dagger$~\cite{liu2023aerialvln}
      & 90.2 & 7.2 & 15.7 & 2.4 \\
    STMR$^\dagger$~\cite{gao2024stmr}
      & 96.3 & 12.6 & 31.6 & 2.7 \\
    CityNavAgent$^\dagger$~\cite{citynavagent2025}
      & 80.8 & 13.9 & 30.2 & 5.1 \\
    LookasideVLN$^\dagger$~\cite{lookasidevln2026}
      & 77.1 & 14.7 & 31.2 & 5.4 \\
    \midrule
    NaVid~\cite{zhang2024navid}
      & 105.1 & 6.8 & 15.5 & 1.1 \\
    OpenFly-Agent~\cite{gao2025openfly}
      & 127.2 & 8.1 & 21.8 & 1.6 \\
    AeroAct~\cite{aeroact2025}
      & 79.6 & 11.4 & 37.7 & 6.3 \\
    \midrule
    \method{}
     
      & \textbf{65.61}
      & \textbf{25.83}
      & \textbf{41.74}
      & \textbf{22.76} \\
    \bottomrule
  \end{tabular*}
  \par\vspace{3pt}
  \begin{minipage}{\columnwidth}
    \scriptsize
    $\dagger$: additional depth, panoramic observations, or odometry.
  \end{minipage}
\end{table}

\textbf{OpenFly.}
\method{} achieves 50.93\%/36.14\% SR and 44.68\%/30.45\% SPL on
Test-Seen/Test-Unseen. Compared with OpenFly-Agent re-evaluated, \method{} improves SR by 25.86 and 17.08 percentage
points on the two splits, respectively. Compared with DreamFly,
\method{} improves SR by 18.89/6.68 points and SPL by 16.46/6.91
points, while reducing NE by 4.62/3.07\,m on Test-Seen/Test-Unseen.
These results show that \method{} achieves substantially better terminal
success and path efficiency than the evaluated 7B language-backbone
methods despite using a 0.5B language backbone.

\textbf{AerialVLN-S.}
\method{} achieves 65.61\,m NE, 25.83\% SR, 41.74\% OSR, and
22.76\% SDTW on Val-Seen, leading all four metrics in
Table~\ref{tab:main}. Compared with AeroAct, \method{} improves SR
and SDTW by 14.43 and 16.46 percentage points, respectively, while
reducing NE by 13.99\,m. Compared with LookasideVLN, which uses
additional inputs, \method{} improves SR by 11.13 points and SDTW by
17.36 points, with an 11.49\,m lower NE. Overall, \method{} achieves
substantially stronger goal-reaching performance and trajectory quality
on AerialVLN-S using only RGB observations.

\subsection{Ablation Study}

We first compare different historical compression methods. We then fix
the proposed learned single-token history and evaluate different
current-view compression and instruction-conditioning designs. Finally,
we study the effect of history capacity and compare with the
single-frame Base policy. All ablations are conducted on OpenFly and
report SR/SPL.

\begin{table}[t]
\caption{Comparison of historical compression methods on OpenFly (SR/SPL in \%).
The current observation remains uncompressed at 256 tokens and
$H=16$.}
\label{tab:history_compression}
\centering
\footnotesize
\setlength{\tabcolsep}{3pt}
\begin{tabular*}{\columnwidth}{@{\extracolsep{\fill}}lcccc@{}}
\toprule
\multirow{2}{*}{History compression} &
\multicolumn{2}{c}{Test-Seen} &
\multicolumn{2}{c}{Test-Unseen} \\
\cmidrule(lr){2-3}\cmidrule(lr){4-5}
& SR$\uparrow$ & SPL$\uparrow$
& SR$\uparrow$ & SPL$\uparrow$ \\
\midrule
MeanPool-1T
& 45.98 & 40.27
& \textbf{36.14} & \textbf{29.53} \\
Learned-1T (Ours)
& \textbf{51.58} & \textbf{45.58}
& 34.90 & 29.36 \\
\bottomrule
\end{tabular*}
\end{table}

\textbf{Historical compression.}
Table~\ref{tab:history_compression} compares two methods for compressing
each historical frame into one token. MeanPool-1T directly averages the
visual features of each historical frame, whereas Learned-1T uses our
learned historical tokenizer. With the same history length $H=16$,
Learned-1T improves Test-Seen SR/SPL from 45.98/40.27\% to
51.58/45.58\%. On Test-Unseen, the two methods perform similarly,
with MeanPool-1T obtaining 36.14/29.53\% and Learned-1T obtaining
34.90/29.36\%. Thus, the learned historical compression provides a
clear advantage on Test-Seen while maintaining comparable performance
on Test-Unseen.

\begin{table}[t]
\caption{Ablation of current-view compression and instruction
conditioning on OpenFly (SR/SPL in \%). All variants use Learned-1T history with
$H=16$.}
\label{tab:current_ablation}
\centering
\footnotesize
\setlength{\tabcolsep}{2.5pt}
\begin{tabular*}{\columnwidth}{@{\extracolsep{\fill}}lcccc@{}}
\toprule
\multirow{2}{*}{Configuration} &
\multicolumn{2}{c}{Test-Seen} &
\multicolumn{2}{c}{Test-Unseen} \\
\cmidrule(lr){2-3}\cmidrule(lr){4-5}
& SR$\uparrow$ & SPL$\uparrow$
& SR$\uparrow$ & SPL$\uparrow$ \\
\midrule

\multicolumn{5}{l}{\scriptsize\itshape Current-view compression} \\
AvgPool-32
& 30.68 & 25.95
& 20.30 & 16.12 \\
GlobalMHA-32
& 47.13 & 41.00
& 34.41 & 28.15 \\

\midrule
\multicolumn{5}{l}{\scriptsize\itshape Instruction conditioning for local aggregation} \\
Local-32, instruction-only
& 48.06 & 42.38
& 31.93 & 25.93 \\
Local-32, history-selected
& 48.78 & 43.03
& 34.65 & 27.98 \\
\textbf{\method{}}
& \textbf{50.93} & \textbf{44.68}
& \textbf{36.14} & \textbf{30.45} \\

\midrule
\multicolumn{5}{l}{\scriptsize\itshape Current-token budget} \\
Local-64, instruction-only
& 47.84 & 42.07
& 34.16 & 28.32 \\
\bottomrule
\end{tabular*}

\par\smallskip
\begin{minipage}{\columnwidth}
\scriptsize
AvgPool-32 directly averages local visual features.
GlobalMHA-32 compresses all 256 current-view tokens into 32 tokens
using global multi-head attention.
For local aggregation, \textit{instruction-only} uses the global
instruction representation, while \textit{history-selected} first uses
the historical representation to select relevant instruction content.
\method{} combines the global and history-selected instruction
representations to guide local aggregation.
\end{minipage}
\end{table}

\textbf{Current-view compression.}
Table~\ref{tab:current_ablation} compares alternative methods for compressing
the 256-token current observation to 32 tokens while using the same Learned-1T
history. AvgPool-32 obtains 30.68/25.95\% SR/SPL on Test-Seen and
20.30/16.12\% on Test-Unseen. GlobalMHA-32 substantially improves
these results to 47.13/41.00\% and 34.41/28.15\%, respectively.
The instruction-only local aggregation achieves 48.06/42.38\% on
Test-Seen and 31.93/25.93\% on Test-Unseen. These results show that
learned aggregation is substantially more effective than direct average
pooling, while the global and local learned aggregation schemes exhibit
different behavior across Seen and Unseen environments.

\textbf{Instruction conditioning for local aggregation.}
We next compare three ways of constructing the instruction representation
used by the same 32-token local aggregation. \textit{Local-32,
instruction-only} uses the global instruction representation without
history-based instruction selection. \textit{Local-32,
history-selected} uses the historical representation to select relevant
instruction content before local aggregation. This improves SR/SPL from
48.06/42.38\% to 48.78/43.03\% on Test-Seen and from
31.93/25.93\% to 34.65/27.98\% on Test-Unseen. The full \method{}
further combines the global and history-selected instruction
representations, reaching 50.93/44.68\% on Test-Seen and
36.14/30.45\% on Test-Unseen. The consistent gains on both splits
support using both global instruction semantics and history-selected
instruction information for current-view aggregation.

\textbf{Current-token budget.}
We compare the instruction-only local aggregation with 32 and 64
current tokens. Reducing the current-token budget from 64 to 32
slightly improves Test-Seen SR/SPL from 47.84/42.07\% to
48.06/42.38\%, while Test-Unseen changes from 34.16/28.32\% to
31.93/25.93\%. Overall, the 32-token setting maintains competitive
navigation performance with a 50\% smaller current-view token budget,
and is therefore adopted in \method{}.

\begin{table}[t]
\caption{Single-frame Base references and history-capacity ablation
on OpenFly. Each entry reports SR/SPL (\%).}
\label{tab:history_ablation}
\centering
\footnotesize
\setlength{\tabcolsep}{1.5pt}
\begin{tabular*}{\columnwidth}{@{\extracolsep{\fill}}lrrrcc@{}}
\toprule
Configuration & $H$ &
\makecell{Current\\tokens} &
\makecell{Max. obs.\\tokens} &
\makecell{Test-Seen\\SR/SPL} &
\makecell{Test-Unseen\\SR/SPL} \\
\midrule

\multicolumn{6}{l}{\scriptsize\itshape Single-frame Base references} \\
Base-Dense
& 0 & 256 & 256
& 42.53/37.31
& 34.41/29.10 \\
Base-AvgPool32
& 0 & 32 & 32
& 37.86/33.34
& 35.39/30.22 \\

\midrule
\multicolumn{6}{l}{\scriptsize\itshape History capacity} \\
\method{} ($H=8$)
& 8 & 32 & 40
& 49.78/43.87
& 33.91/27.70 \\
\textbf{\method{} ($H=16$)}
& 16 & 32 & 48
& \textbf{50.93/44.68}
& \textbf{36.14/30.45} \\
\method{} ($H=31$)
& 31 & 32 & 63
& 49.78/44.13
& 34.16/28.26 \\
\bottomrule
\end{tabular*}
\end{table}

\textbf{History capacity.}
Table~\ref{tab:history_ablation} summarizes the single-frame references and
history capacities. Base-Dense is the single-frame Base policy with the original 256-token
current observation, while Base-AvgPool32 replaces the current
observation with 32 average-pooled tokens. They provide references
without historical input. For \method{}, $H=8$, 16, and 31 correspond
to a maximum of 40, 48, and 63 observation-derived tokens,
respectively. Each history capacity is trained separately in Stage~2
from the same Base checkpoint and Stage~1 cache. Among the three
history settings, $H=16$ achieves the highest SR/SPL on both splits:
50.93/44.68\% on Test-Seen and 36.14/30.45\% on Test-Unseen.
Increasing the history capacity to $H=31$ does not provide further
improvement.

\subsection{Efficiency and Closed-Loop Evaluation}
\label{sec:onboard}

\begin{figure}[t]
  \centering
  \includegraphics[width=\columnwidth]{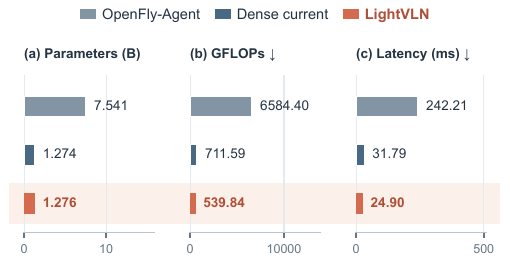}
  \caption{Model size, computation, and inference latency.}
  \label{fig:full_efficiency_plot}
\end{figure}

\textbf{Offline efficiency.}
Dense current and \method{} use the same learned single-token history;
they differ only in the current-view representation.
\method{} reduces the current
visual tokens from 256 to 32 and the total observation-derived tokens
from 272 to 48, corresponding to an 82.35\% reduction.
Figure~\ref{fig:full_efficiency_plot} shows that full-model computation
decreases from 711.59 to 539.84 GFLOPs, while RTX 4090 latency decreases
from 31.79 to 24.90\,ms, corresponding to reductions of 24.14\% and
21.67\%, respectively. The two models have similar parameter counts
(1.274B vs.\ 1.276B). Because the visual encoders remain
unchanged, the reduction in full-model computation is smaller than the
82.35\% reduction in observation-derived tokens. For reference,
OpenFly-Agent contains 7.541B parameters and requires 242.21\,ms per
inference under the same profiling setup.

\begin{figure}[t]
  \centering
  \includegraphics[width=0.8\columnwidth]{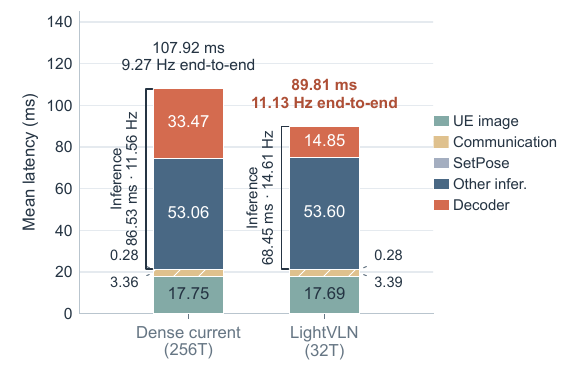}
  \caption{Latency breakdown for Dense current and \method{}. Stacked
  bars separate image acquisition, communication, pose updates, and
  model inference.}
  \label{fig:orin_power_plot}
\end{figure}

\textbf{Onboard efficiency.}
In the closed-loop onboard-compute HIL evaluation, \method{} averages 68.45\,ms per model
inference (14.61\,Hz) and 89.81\,ms end to end (11.13\,Hz).
Dense current requires 86.53\,ms (11.56\,Hz) and 107.92\,ms
(9.27\,Hz), respectively (Fig.~\ref{fig:orin_power_plot}).
Compression reduces model latency by 20.89\% and end-to-end latency
by 16.78\%; LLM decoder latency decreases from 33.47 to 14.85\,ms.
For \method{} and Dense current, respectively, image acquisition
averages 17.69/17.75\,ms and communication overhead 3.39/3.36\,ms.
SetPose takes 0.28\,ms for each variant. The reduction therefore
comes primarily from model inference within the closed loop.

\textbf{Campus navigation results.}
On the 500-episode campus unseen evaluation set, \method{} obtains
8.00\% SR, 65.00\% OSR, and 56.06\,m NE. The large OSR--SR gap
indicates that reaching the goal vicinity does not always translate
into terminal success in the unseen environment. These results assess
closed-loop navigation transfer with physical onboard inference in
the reconstructed environment.

\begin{figure*}[t]
  \centering
  \includegraphics[width=0.9\textwidth]{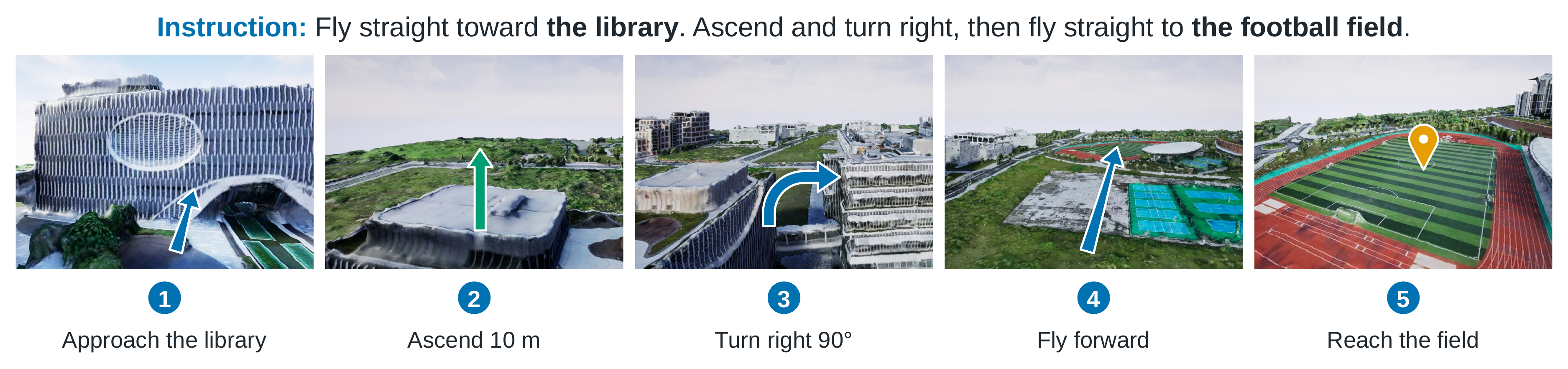}
  \caption{A successful \method{} rollout under closed-loop onboard-compute HIL
  in the campus environment, visualized with representative observations
  and grouped predicted actions.}
  \label{fig:qualitative-route}
\end{figure*}

\textbf{Qualitative closed-loop example.}
Figure~\ref{fig:qualitative-route} shows a successful \method{} rollout
in the campus environment under closed-loop onboard-compute HIL,
with representative observations and
grouped predicted action segments. 

\section{Conclusion}
We presented \method{}, a lightweight history-aware aerial VLN framework
built on a compact 0.5B language backbone with compact historical and
current visual representations. \method{} compresses each historical
frame into a single token by reusing visual features already computed
by the policy, and uses historical information to guide local aggregation
of the current observation from 256 to 32 tokens. With up to 16
historical frames, the policy uses at most 48 observation-derived tokens.
\method{} achieves 50.93\%/36.14\% SR on public OpenFly
Test-Seen/Test-Unseen and 25.83\% SR on AerialVLN-S Val-Seen.
Relative to the dense-current counterpart, the compact representation
reduces full-model computation by 24.14\% and RTX 4090 inference
latency by 21.67\%. On Jetson Orin NX 16\,GB, \method{} reaches
14.61\,Hz model inference and 11.13\,Hz end-to-end decision updates
in the closed-loop onboard-compute HIL system. Future work will further improve
cross-environment robustness.

\bibliographystyle{IEEEtran}
\bibliography{ref}
\end{document}